%% file: acl_latex.tex
\pdfoutput=1
\documentclass[11pt]{article}
\usepackage[]{acl}
\usepackage{times}
\usepackage{latexsym}
\usepackage{amsfonts} 
\usepackage{etoolbox} 
\usepackage{booktabs} 
\usepackage[noend]{algpseudocode} 
\usepackage{algorithmicx,algorithm} 
\usepackage{multirow} 
\usepackage{amsmath} 
\usepackage{bbding} 
\usepackage{caption} 
\usepackage{subfigure} 
\usepackage{soul} 
\usepackage{url} 
\usepackage{hyperref} 
\usepackage{graphicx} 
\usepackage[bottom]{footmisc} 
\usepackage{makecell} 
\usepackage{float} 
\usepackage{amssymb}

\definecolor{mycolor1}{RGB}{235, 83, 83} 
\definecolor{mycolor2}{RGB}{255, 191, 2}
\definecolor{mycolor3}{RGB}{158, 158, 158} 
\definecolor{mycolor4}{RGB}{163, 26, 203} 
\definecolor{mycolor5}{RGB}{255, 149, 81} 
\definecolor{mycolor6}{RGB}{54, 174, 124}
\definecolor{mycolor7}{RGB}{0, 158, 255} 
\definecolor{mycolor8}{RGB}{173, 105, 176}
\definecolor{mycolor9}{RGB}{60, 92, 160}
\definecolor{mycolor10}{RGB}{211, 111, 65}
\definecolor{mycolor11}{RGB}{71, 155, 86}
\definecolor{mycolor12}{RGB}{181, 57, 65}
\colorlet{soulyellow}{yellow!50}
\colorlet{soulorange}{orange!30}
\colorlet{soulblue}{cyan!20}
\colorlet{soulgreen}{green!20}

\usepackage{xspace}

\newcommand{\method}{{\textsc{DiffusEmp}}\xspace}
\newcommand{\dataset}{\textsc{EmpatheticDialogue}\xspace}
\newcommand{\cs}{control signals\xspace}
\usepackage[T1]{fontenc}
\usepackage[utf8]{inputenc}
\usepackage{microtype}

\title{\method: A Diffusion Model-Based Framework with \\ Multi-Grained Control for Empathetic Response Generation}

\author{
  Guanqun Bi${}^{1,2}$,
  Lei Shen${}^{3}$,
  Yanan Cao${}^{1,2}$\thanks{~ Corresponding authors.},
  Meng Chen${}^{3}$\footnotemark[1],
  \\
  {\bf Yuqiang Xie${}^{1,2}$},
  {\bf Zheng Lin${}^{1,2}$},
  {\bf Xiaodong He${}^{3}$}
  \\
  ${}^{1}$Institute of Information Engineering, Chinese Academy of Sciences, Beijing, China \\
  ${}^{2}$School of Cyber Security, University of Chinese Academy of Sciences, Beijing, China \\
  ${}^{3}$JD AI Research, Beijing, China \\
  {\tt \{biguanqun,caoyanan,xieyuqiang,linzheng\}@iie.ac.cn} \\
  {\tt \{shenlei20,chenmeng20,xiaodong.he\}@jd.com} \\
}

\begin{document}
\maketitle
\begin{abstract}

Empathy is a crucial factor in open-domain conversations, which naturally shows one’s caring and understanding to others. 
Though several methods have been proposed to generate empathetic responses, existing works often lead to monotonous empathy that refers to generic and safe expressions.
In this paper, we propose to use explicit control to guide the empathy expression and design a framework \method based on conditional diffusion language model to unify the utilization of dialogue context and attribute-oriented control signals.
Specifically, \textit{communication mechanism}, \textit{intent}, and \textit{semantic frame} are imported as multi-grained signals that control the empathy realization from coarse to fine levels. 
We then design a specific masking strategy to reflect the relationship between multi-grained signals and response tokens, and integrate it into the diffusion model to influence the generative process.
Experimental results on a benchmark dataset \dataset show that our framework outperforms competitive baselines in terms of controllability, informativeness, and diversity without the loss of context-relatedness.

\end{abstract}

\input{Tex/1_introduction.tex}
\input{Tex/2_related_work.tex}
\input{Tex/3_method.tex}
\input{Tex/4_experiments.tex}
\input{Tex/5_discussion.tex}

\input{Tex/6_conclusion.tex}

\section*{Acknowledgement}
We thank the reviewers for their detailed and insightful advice. 
This work is supported by the National Key Research and Development Program of China (NO.2022YFB3102200) and Strategic Priority Research Program of the Chinese Academy of Sciences with No. XDC02030400.

\input{Tex/7_limitations.tex}
\input{Tex/8_ethics.tex}

\bibliography{anthology,custom}
\bibliographystyle{acl_natbib}

\clearpage

\appendix

\input{Tex/a1_exp_details.tex}
\input{Tex/a2_future.tex}
\input{Tex/a3_cases.tex}

\end{document}

%% file: Tex/1_introduction.tex
\section{Introduction}
\begin{figure}
    \centering
    \includegraphics[width=0.48\textwidth]{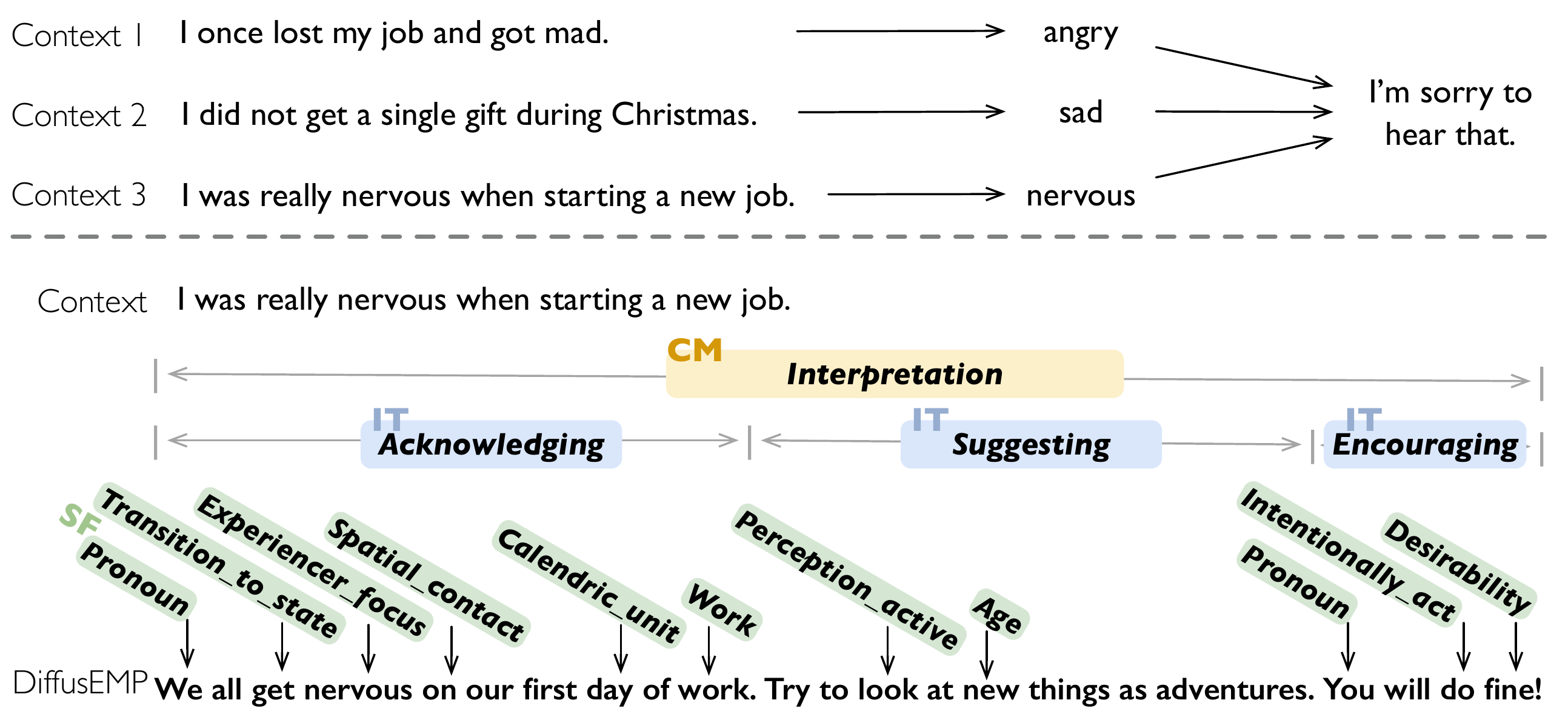}
    \caption{A monotonous empathetic response (upper) and an informative empathetic response (lower). ``CM'', ``IT'', and ``SF'' are abbreviations for ``Communication Mechanism'', ``Intent'', and ``Semantic Frame'', which represent control signals at the utterance, sentence, and token level respectively. }
    \label{fig-intro}
\end{figure}

Empathetic response generation, as a conditional text generation task, aims to endow agents with the ability to understand interlocutors and accurately express empathy in their communication \cite{rashkin-etal-2019-towards,lin-etal-2019-moel,li-etal-2020-empdg,shen2021constructing}. 
However, the generated responses tend to be generic and monotonous \cite{Chen2022EmpHiGE}, i.e.,  showing shallow empathy and few connections to the context. As shown in the upper part of \figureautorefname~\ref{fig-intro}, ``I'm sorry to hear that.'' is used as a reaction to different contexts with negative feelings.
To alleviate the problem, existing works mainly incorporate emotion or knowledge modules into the encoder-decoder framework and train their models with the maximum likelihood estimation (MLE) \cite{rashkin-etal-2019-towards,lin-etal-2019-moel,majumder-etal-2020-mime,li-etal-2020-empdg,CEM2021,li-etal-2022-kemp}.

Recently, diffusion models \cite{Ho2020DenoisingDP,Dhariwal2021DiffusionMB} have emerged as a brand-new and promising paradigm for generative models.  
A few prior works that explored using diffusion models on text data are mainly designed for unconditional text generation \cite{Austin2021StructuredDD,Hoogeboom2021ArgmaxFA,he2022diffusionbert}.
For text generation with extra conditions (control signals or contexts), Diffusion-LM \cite{Li-2022-DiffusionLM} applies extra-trained classifiers to make the generated text satisfy input signals like sentiment and syntactic structure.
DiffuSeq \cite{gong2022diffuseq} is proposed as a classifier-free diffusion model that uses ``partial noising'' in the forward process to distinguish the input and output text.

In this paper, we add control signals to empathetic response generation and propose a diffusion model-based framework, \method, to solve the aforementioned monotonous empathy problem. 
First, since empathy is a multi-dimensional factor \cite{Davis1980AMA}, i.e., several factors affect the realization of empathy, we use explicit control signers at different levels to guide response generation.
At the utterance level, \textit{communication mechanism} (CM) \cite{sharma-etal-2020-computational} divides text-based empathy into emotional reaction, interpretation, and exploration to describe the high-level functionality. Then, we use \textit{intent} (IT) \cite{welivita-pu-2020-taxonomy} to reflect the behaviors of an agent in each sentence\footnotemark[2]\footnotetext[2]{An utterance, the response here, may consist of more than one sentence.}, such as questioning (e.g., \textit{What happened to you?}). Finally, the fine-grained signal \textit{semantic frame} (SF) \cite{baker-etal-1998-berkeley-framenet} is imposed on each token, which represents their universal categories of events, concepts, and relationships. An example of how multi-grained control signals work is illustrated in the lower part of \figureautorefname~\ref{fig-intro}. To have exact guidance over responses, these signals are extracted from golden responses in the training process, while during inference, an emotion-enhanced matching method is used to obtain response candidates as the source of control signals.

We then design a diffusion model to make the generated responses not only relevant to dialogue contexts but also express specific empathy under the multi-grained control.
The dialogue context, multi-grained control, and response are considered as the model input. 
For the forward diffusion process, we apply the partial noising \cite{gong2022diffuseq} strategy so that both the context and control signals are unchanged, and only the response is noised.
To fulfill the reverse diffusion process, we use the transformer architecture \cite{vaswani2017attention} and introduce a masking strategy to indicate the control range of each signal on response tokens.
Specifically, each CM/IT controls all tokens in an utterance/sentence, while an SF term corresponds to exactly one token. 
Tokens out of the control range are masked in the self-attention layer.
Finally, we conduct experiments on a benchmark dataset \dataset to demonstrate the effectiveness of \method.

The main contribution of this paper is threefold:
(1) We introduce explicit multi-grained control signals to solve the monotonous empathy problem, and convert the empathetic response generation into a controllable setting. 
(2) We propose \method, a novel diffusion model-based framework, to unify the utilization of dialogue context and control signals, achieve elaborate control with a specific masking strategy, and integrate an emotion-enhanced matching method to produce diverse responses for a given context.
(3) Experimental results show that our method outperforms competitive baselines in generating informative and empathetic responses.

%% file: Tex/2_related_work.tex
\section{Related Work}

\subsection{Empathetic Response Generation}
\citet{rashkin-etal-2019-towards} firstly formulate the empathetic response generation task and construct the \dataset dataset.
Existing works that focus on this task can be divided into two lines.
The first is to detect and utilize the user's emotion with diverse structures \cite{lin-etal-2019-moel,majumder-etal-2020-mime,shen2021constructing}.
The second is to consider cognition-based factors other than emotions (EM), such as dialogue act (DA) \cite{welivita-pu-2020-taxonomy}, communication mechanism (CM) \cite{sharma-etal-2020-computational}, emotion cause \cite{jiang2019improving}, psychological skill \cite{kim-etal-2021-perspective}, and commonsense \cite{Sabour2021CEM, li-etal-2022-kemp}.
\citet{Zheng2021CoMAEAM} propose a framework CoMAE to model the relationship among CM, DA, and EM at the utterance level.
The differences between CoMAE and \method are: 
(1) Instead of predicting each factor based on the context representation, \method explicitly uses control signals that are highly related to a response as task input. 
(2) We achieve the elaborate control with multi-grained signals, i.e., tokens in response are influenced by different signals, while CoMAE applies the same combined factor to all decoding positions.

\begin{figure*}[htbp]
    \centering
    \includegraphics[width=1.0\textwidth]{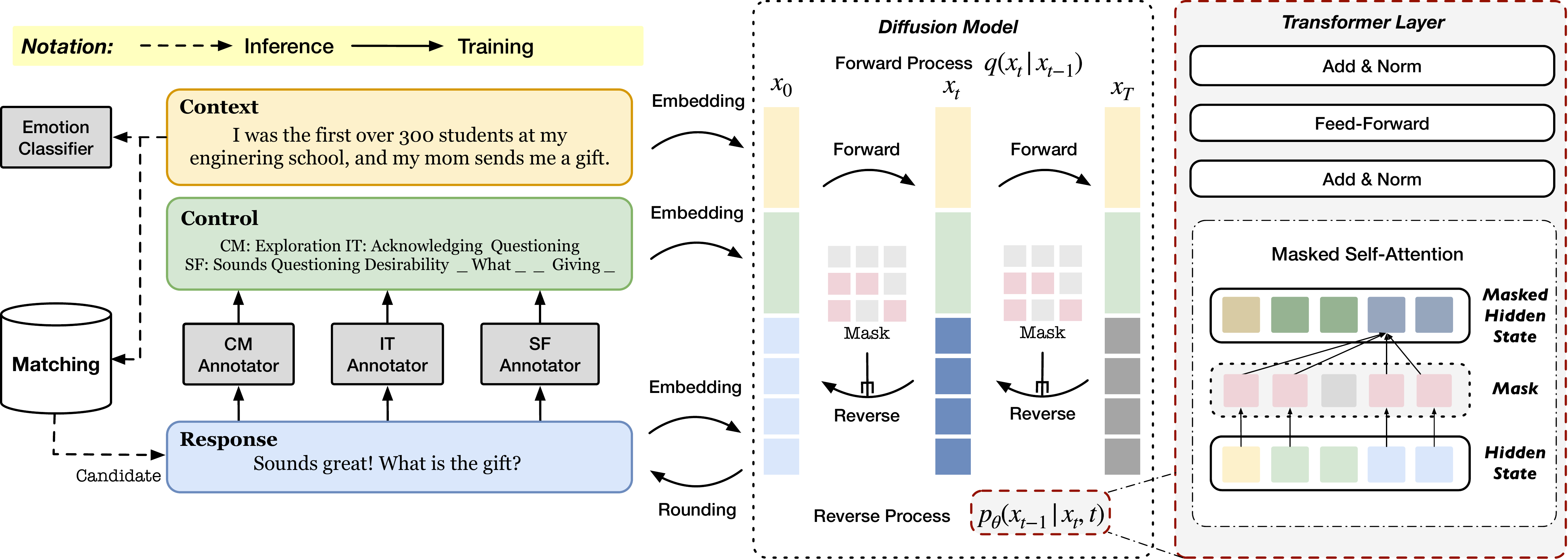}
    \caption{The overview of \method. The left part describes the training and inference stages, the middle part shows the forward process and reverse process in the diffusion model, and the right part illustrates details in a Transformer \cite{vaswani2017attention} block with control-range masking for the reverse process.}
    \label{fig-overview}
\end{figure*}

\subsection{Diffusion Models}

Diffusion models are a class of generative models with promising performance and have been used in a variety of real-world applications.
Most existing works of diffusion models focus on continuous data, such as vision \cite{Nichol2021GLIDETP, Radford2021LearningTV, rombach2021highresolution} and audio \cite{Popov2021GradTTSAD, Yang2022DiffsoundDD, Tae2021EdiTTSSE}.
Due to the discrete nature of text data, the utilization of diffusion models for NLP is challenging.
\citet{Hoogeboom2021ArgmaxFA} and \citet{Austin2021StructuredDD} extend diffusion models to discrete state spaces for character-level text generation. 
Diffusion-LM \cite{Li-2022-DiffusionLM} uses embedding and rounding strategy to bridge the continuous and discrete domain, and trains extra classifiers for controllable text generation.
DiffuSeq \cite{gong2022diffuseq} leverages partial noising for sequence-to-sequence text generation to keep the text input unchanged in the forward process.
DiffusionBERT \cite{he2022diffusionbert} combines pretrained language models with absorbing-state discrete diffusion models for text.
To the best of our knowledge, we are the first to achieve controllable empathetic response generation using a diffusion model.

%% file: Tex/3_method.tex
\section{\method}

In this paper, we perform empathetic response generation in a controllable setting. 
The dialogue context is an alternating sequence of utterances from a speaker and a listener, i.e. $\mathbf{w}^u=\{u_1,u_2,\dots,u_{n}\}$.
Here, we aim to generate an empathetic and context-related response $\mathbf{w}^y=\{y_1,y_2,\dots,y_{n}\}$ conditioned on the given context $\mathbf{w}^u$ and a set of control signals $\mathbf{w}^c$ obtained in advance (Section \ref{sec-cf}).
Then, the context, control signals, and response are concatenated and fed into a diffusion model with control-range masking (Section \ref{sec-diffmodel}). 
In the training process, golden responses are used to extract control signals, while during inference, we integrate an emotion-enhanced matching method to get proper response candidates (Section \ref{sec-train/inf}).
The framework of \method is illustrated in \figureautorefname~\ref{fig-overview}.

\subsection{Acquisition of Control Signals}
\label{sec-cf}

To better model and express multi-dimensional empathy, we use control signals at different levels. 
However, the benchmark dataset \dataset does not contain such annotations.
Here, we introduce three types of signals used in this paper and the way to collect them for each golden response or response candidate using pre-trained tagging models.
The definition and components of empathy in psychology are complex\cite{Davis1980AMA, Waal2008PuttingTA, Decety2008FromER}, and we choose the control signals that intersect with computational linguistics.
Note that the design of \method is not limited to the following control signals, other factors of empathy can also be used.

\noindent\textbf{Communication Mechanism (CM).} 
We employ the taxonomy in \citet{sharma-etal-2020-computational}: \textit{Emotional Reaction (ER), Interpretation (IP),} and \textit{Exploration (EX)}. 
ER expresses emotions such as warmth, compassion, and concern,
IP represents an understanding of feelings and experiences inferred from the speaker,
and EX stands for exploring the feelings and experiences not stated in previous utterances. 
Following \citet{sharma-etal-2020-computational}, we use three RoBERTa-based \cite{liu2019roberta} classifiers to individually identify whether a response implies a certain mechanism. 

\noindent\textbf{Intent (IT).}
A previous analysis \cite{welivita-pu-2020-taxonomy} argues that humans demonstrate a wide range of intents when regulating empathy and proposes a dataset \textsc{EmpatheticIntent}. 
Besides, many works \cite{xie-etal-2022-comma,Zheng2021CoMAEAM} insist that intents and emotions have a strong relationship.
Specifically, listeners are much more likely to respond to positive or negative emotions with specific empathetic intents such as \textit{acknowledgment}, \textit{consolation}, and \textit{encouragement}, rather than only expressing similar or opposite emotions.
We train a BERT-based \cite{Devlin2019BERTPO} classifier on \textsc{EmpatheticIntent} to label responses.

\noindent\textbf{Semantic Frame (SF).}
Semantic frames are based on FrameNet \cite{baker-etal-1998-berkeley-framenet}, a linguistic knowledge graph containing information about lexical and predicate-argument semantics. 
The frame of a token represents its universal categories of events, concepts, and relationships, and can be regarded as a high-level abstraction of meaning. 
For example, tokens like \textit{bird, cat, dog, horse, sheep} share the same frame label \textit{Animals}.
Here, we utilize the open-SESAME model \cite{Swayamdipta2017FrameSemanticPW} to extract semantic frames from responses.

\begin{table}[]
    \centering\small
    \resizebox{0.35\textwidth}{!}{%
    \begin{tabular}{c|ccc}
        \hline
        
        \hline
        \textbf{Signal Type}& \textbf{Accuracy} & \textbf{F1}    & \textbf{\#Classes} \\ \hline\hline
        \textbf{CM-ER} & 79.43    & 74.46 & 2         \\
        \textbf{CM-IP} & 84.04    & 62.60 & 2         \\
        \textbf{CM-EX} & 92.61    & 72.58 & 2         \\
        \textbf{IT}    & 87.75    & 87.71 & 9         \\
        \textbf{SF}    & -       & 86.55 & 1222      \\ 
        \hline
        
        \hline
    \end{tabular}
}
\caption{The performance of tagging tools used to get control signals. Since SF is from a frame semantic parsing task, we only report the F1 score following the original task setting.}
\label{table-classifier}
\end{table}

The performance of tagging tools is listed in \tableautorefname~\ref{table-classifier}.
Note that control signal tokens are concatenated into a flat sequence from coarse to fine.

\subsection{Diffusion Model with Control-Range Masking}
\label{sec-diffmodel}

A diffusion model contains a forward process and a reverse process. 
We first concatenate a context with the control signals and corresponding response, i.e., $\mathbf{w}=\mathbf{w}^u\oplus\mathbf{w}^c\oplus\mathbf{w}^y$.
Then we use an \textit{embedding} function \cite{Li-2022-DiffusionLM} $\textsc{Emb}(\mathbf{\cdot})$ to map the discrete text $\mathbf{w}$ into a continuous representation $\mathbf{x}_0=\mathbf{u}_0\oplus\mathbf{c}_0\oplus\mathbf{y}_0$, where $\mathbf{u}_0$, $\mathbf{c}_0$, and $\mathbf{y}_0$ represent parts of $\mathbf{x}_0$ that belong to $\mathbf{w}^u$, $\mathbf{w}^c$, and $\mathbf{w}^y$, respectively.

\noindent\textbf{Forward Process.}
In forward process $q$, the model adds noise to the original sample $\mathbf{x}_0$ step by step:
\begin{equation}
    q(\mathbf{x}_t\vert\mathbf{x}_{t-1})=\mathcal{N}(\mathbf{x}_t;\sqrt{1-\beta_t}\mathbf{x}_{t-1},\beta_t\mathbf{I}),
\end{equation}
where $\mathbf{x}_1,...,\mathbf{x}_T$ make up a chain of Markov variants and $\mathbf{x}_T \sim \mathcal{N}(0, \mathbf{I})$.
$\beta_t\in(0,1)$ is a noise schedule that controls the noise scale added in each step.
Note that the conventional diffusion models corrupt the entire $\mathbf{x}_0$.
However, empathetic response generation is a conditional text generation (Seq2Seq) task and we only concern with the generative effect on response.
Therefore, we use partial noising \cite{gong2022diffuseq} to only impose noise on the parts of $\mathbf{x}_t$ that belong to $\mathbf{w}^y$, i.e., $\mathbf{y}_t$.

\noindent\textbf{Reverse process.}
Once the forward process is completed, the reverse process aims to gradually recover $\mathbf{x}_0$ by denoising $\mathbf{x}_T$ according to:
\begin{equation}
    p_\theta(\mathbf{x}_{t-1}\vert \mathbf{x}_t, t)=\mathcal{N}(\mathbf{x}_{t-1};\mu_\theta(\mathbf{x}_t,t),\sigma_\theta(\mathbf{x}_t,t)),
\end{equation}
where $\mu_\theta(\cdot)$ and $\sigma_\theta(\cdot)$ are predicted mean and standard variation of $q(\mathbf{x}_{t-1}|\mathbf{x}_t)$ (derived using Bayes' rule) in forward process and can be implemented by a Transformer \cite{vaswani2017attention} model $f_\theta$.
In the reverse process, we add a \textit{rounding} step \cite{Li-2022-DiffusionLM}, parameterized by $p_\theta(\mathbf{w}|\mathbf{x}_0)=\prod^n_{i=1}p_\theta(w_i|x_i)$, where $p_\theta(w_i|x_i)$ is a softmax distribution.

\begin{figure}
    \centering
    \includegraphics[width=0.48\textwidth]{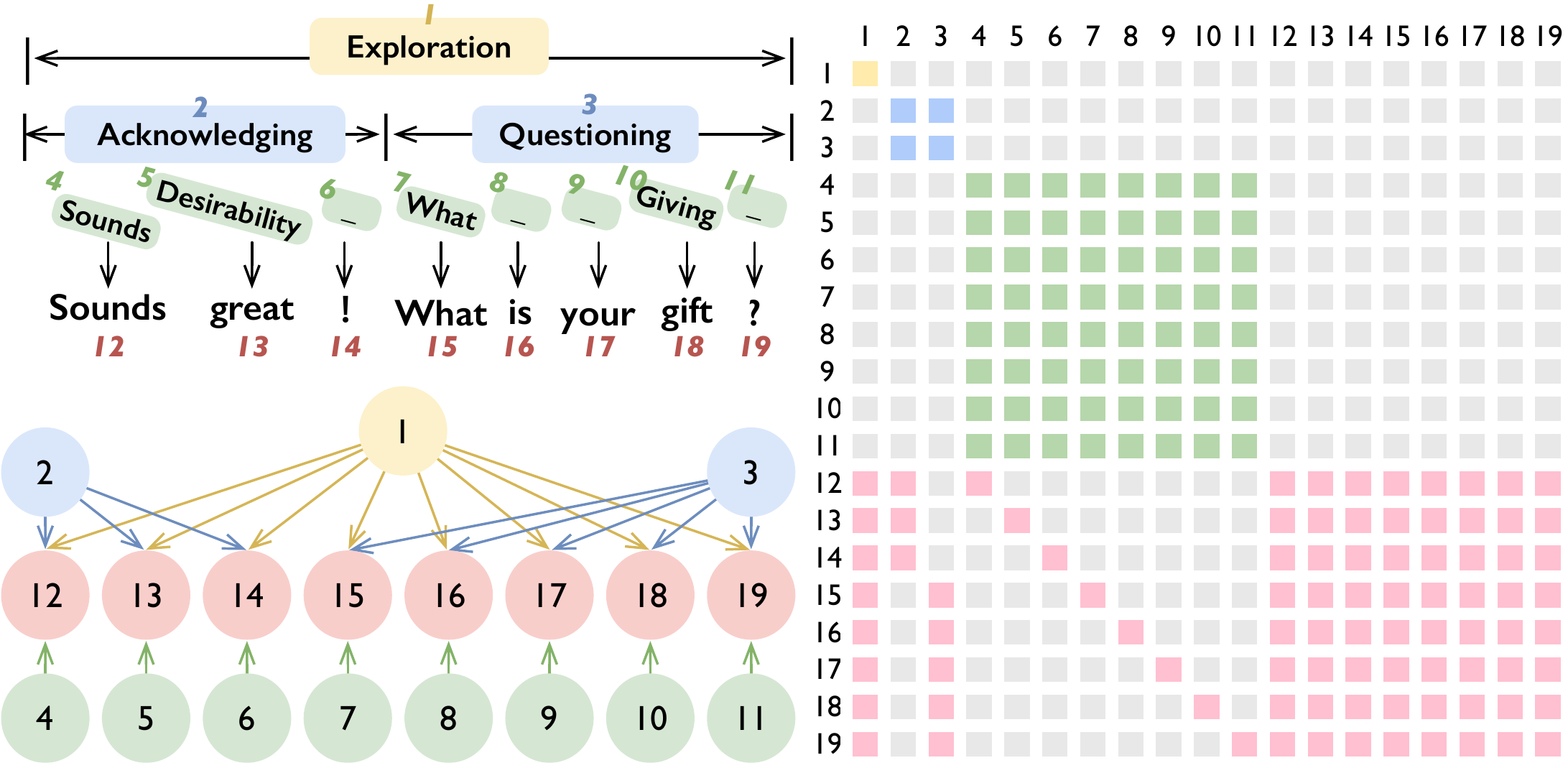}
    \caption{An example of \cs and control-range masking. The upper left part shows a response with labeled signals, the lower left part illustrates the control range of each signal on response tokens, and the right part is the corresponding mask matrix. ``-'' means the SF signal is empty.}
    \label{fig-control-example}
\end{figure}

\noindent\textbf{Control-Range Masking.}
The non-autoregressive nature of conventional diffusion models make one input token can attend to all other tokens with the full self-attention mechanism to update its representation.
Instead, we need to distinguish between tokens of control signals and responses, and further model the relationship between them with a mask matrix $M$ and integrate it into the self-attention layer in Transformer:
\begin{align}
    &Q^{i+1},K^{i+1},V^{i+1}=h^iW_q,h^iW_k,h^iW_v,\\
    &S^{i+1}=softmax(\frac{Q^{i+1}K^{i+1 \mathsf{T}}+M}{\sqrt{d_k}}),\\
    &h^{i+1}=S^{i+1}V^{i+1},
\end{align}
where $W_q, W_k$ and $W_v$ are trainable parameters, $h^i$ is the hidden state of the $i$-th transformer layer. $d_k$ is the dimension of $K$, which is used for scaling.

Basically, if token $i$ controls $j$, then the calculation of $j$ is influenced by $i$. In terms of implementation, we do not mask $i$ when updating the representation of $j$.
Particularly, tokens at the same level, including IT signal tokens, SF signal tokens, and response tokens, are also designed to control each other, thus ensuring the overall logic and fluency of the generated responses.
For example, it is reasonable that \textit{Sympathizing} is followed by \textit{Questioning} at the intent level, i.e., expressing more concerns by questioning after showing sympathy for a negative situation or feeling.
Therefore, to model the control relationship among tokens, we design the  control-range masking and utilize it in the self-attention layer of $f_\theta$.
Specifically, for a mask matrix, the value on position $(i, j)$ is 0 if token$_j$ is controlled by token$_i$; otherwise is negative infinity:
\begin{equation}
M(i,j)=\left\{
\begin{aligned}
0,& &  i \Rightarrow j \\
-\inf,& & i \not\Rightarrow j \\
\end{aligned}
\right.
\end{equation}

\figureautorefname~\ref{fig-control-example} gives an example of control-range masking. 
For the intent signal \textit{Acknowledging} (index 2), it is visible to \textit{Questioning} (line 3) and corresponding response tokens \textit{Sounds great!} in the first sentence (line 12-14).
Meanwhile, since the response token \textit{great} (line 13) is controlled by \textit{Exploration} (index 1), \textit{Acknowledge} (index 2), \textit{Desirability} (index 5), and the rest of response tokens (index 12-19), it attends to them in the mask matrix.

With the existence of control-range masking, we can elaborately guide the generation of each response token with signals from different levels that reflect diverse factors for empathy expression.

\subsection{Training and Inference}
\label{sec-train/inf}
\noindent\textbf{Training.}
In the training process, we label control signals based on golden responses as described in \ref{sec-cf}.
To train model $f_\theta$ in the reverse process, we minimize the variational lower bound following \citet{gong2022diffuseq}:
\begin{equation}
\begin{split}
\mathcal{L}_{\text{vlb}}= &  \sum_{t=2}^T||\mathbf{y}_0-\tilde f_{\theta}(\mathbf{x}_t, t)||^2 \\ 
& + ||\textsc{Emb}(\mathbf{w}^y)-\tilde f_{\theta}(\mathbf{x}_1, 1)||^2 \\
& + \mathcal{R}(||\mathbf{x}_0||^2),
\end{split}
\end{equation}
where $\tilde f_{\theta}(\mathbf{x}_t, t)$ denotes the fractions of recovered $\mathbf{x_0}$ corresponding to $\mathbf{y}_0$, and $\mathcal{R}(\cdot)$ is a mathematically equivalent regularization term to regularize the
embedding learning.

\noindent\textbf{Inference.}
During inference, since golden responses are unavailable, we design an emotion-enhanced matching method to obtain response candidates and use them to extract control signals.
We treat dialogue contexts in the training set as the candidate pool and use each context in the test set as a query to perform context-context matching.
Then the response corresponding to a returned context with the highest similarity is used as the candidate. 
Regarding the importance of emotions in empathetic response generation, we consider two aspects to score each candidate, semantic similarity and emotional consistency, in context-context matching.
Specifically, we first train a BERT model \cite{Devlin2019BERTPO} on the training set to classify emotions for contexts.
Then, we use this model to get emotional distribution for contexts in both the candidate pool and queries.
Finally, we compute the cosine similarity of both sentence embeddings and predicted emotional distributions for each query-context pair. 
The contexts are re-ranked according to a weighted sum of two similarity scores:
\begin{equation}
    Score=\textsc{Sim}_{\text{semantic}}+\gamma \textsc{Sim}_{\text{emotional}},
    \label{score-match}
\end{equation}
where $\gamma$ is a hyperparameter to balance the semantic and emotional similarity.

%% file: Tex/4_experiments.tex
\section{Experimental Setup}

\subsection{Dataset} 
\dataset \cite{rashkin-etal-2019-towards} dataset comprises 24,850 open-domain multi-turn conversations between two interlocutors.
Each conversation contains one emotion label, a situation where the speaker feels the exact emotion, and utterances about the speaker’s descriptions of the situation or the listener’s empathetic replies. 
There are 32 evenly-distributed emotion labels in the dataset. 
We apply the data provided by the original paper with the split ratio of 8:1:1 for training/validation/test set
and use the script released by \citet{lin-etal-2019-moel} to preprocess the data.

\begin{table*}[htb]
\centering
\resizebox{\textwidth}{!}{%
\begin{tabular}{l|c|c|c|ccc|cccc|c} 
\hline

\hline
\multicolumn{1}{c|}{\multirow{2}{*}{\textbf{Method}}} & \multirow{2}{*}{\textbf{\#Params}} & \multicolumn{2}{c|}{\textbf{Relevance}} & \multicolumn{3}{c|}{\textbf{Controllability}}                      & \multicolumn{4}{c|}{\textbf{Informativeness}}     & \multicolumn{1}{c}{\textbf{Length}}                           \\ 
\cline{3-12}
\multicolumn{1}{c|}{}                                 &                           & \textbf{BERTScore} ↑  & \textbf{MIScore} ↓ & \textbf{ACC-CM} ↑ & \textbf{ACC-IT} ↑ & \textbf{F1-SF} ↑ & \textbf{D1} ↑ & \textbf{D2} ↑  & \textbf{D4} ↑  & \textbf{sBL} ↓ & \textbf{AvgLen} ↑     \\ 
\hline\hline
\multicolumn{10}{l}{\textit{Transformer-Based Methods}} & \multicolumn{1}{l}{}  \\ 
\hline
\textbf{TRS} & 15M & 0.5717 & 4598.26 & 60.98 & 22.07 & 15.74 & 0.42 & 1.55 & 4.26 & 13.63 & 10.53 \\
\textbf{MTRS} & 15M & 0.5735 & 7156.26 & 60.48 & 25.77 & 15.62 & 0.50 & 1.89 & 5.56 & 11.26 & 9.92  \\
\textbf{MoEL}& 21M & 0.5758  & 14595.61 & 59.29 & 26.20 & 16.51 & 0.40 & 1.65 & 4.62 & 12.83 & 11.47 \\
\textbf{MIME} & 17M & 0.5800 & 4878.71 & 61.16 & 22.00 & 16.54 & 0.26 & 0.87 & 2.15 & 14.21 & 11.12 \\
\textbf{EmpDG} & 29M & 0.5745 & 9088.11 & 61.94 & 20.06 & 17.36 & 0.60 & 2.54 & 7.75 & 11.78 & 10.11 \\
\textbf{CEM} & 17M & 0.5713 & 7635.05 & 62.28 & 30.09 & 14.20 & 0.54 & 2.00 & 4.98 & 9.13 & 8.25 \\ 
\hline
\multicolumn{10}{l}{\textit{Pre-Trained Language Model-Based Methods}}  & \multicolumn{1}{l}{}  \\ 
\hline
\textbf{TransferTransfo} & 117M & 0.5634 & 2138.39 & 59.70 & 25.08 & 18.39 & 2.81 & 17.22 & 36.54 & 2.68 & 11.40 \\
\textbf{BART} & 140M & \textbf{0.5977} & 706.31  & 60.39 & 30.69 & 18.98 & \textbf{2.88} & 14.12 & 38.82 & 2.79 & 11.09 \\ 
\hline
\multicolumn{10}{l}{\textit{Diffusion Model-Based Methods}}    & \multicolumn{1}{l}{}  \\ 
\hline
\textbf{DiffuSeq} & 91M & 0.5101 & 715.95  & 59.23 & 28.58 & 17.26 & 1.79  & 26.97 & \textbf{88.17} & 1.29 & 10.30 \\
\textbf{\method} & 91M & 0.5205 & \textbf{626.92}  & \textbf{92.36} & \textbf{84.24} & \textbf{52.79} & 2.84  & \textbf{29.25} & 73.45 & \textbf{1.09} & \textbf{14.12} \\
\hline
\multicolumn{10}{l}{\textit{References}}  & \multicolumn{1}{l}{}  \\ 
\hline
\textbf{\method(Oracle)} & 91M & 0.7458 & 615.13 & 92.38 & 83.66 & 51.95 & 2.84 & 30.46 & 89.35 & 1.11 & 14.01 \\
\textbf{Human} & - & 1.0000 & 507.97 & 100.00 & 100.00 & 98.40 & 19.49 & 43.55 & 49.02 & 0.85 & 13.04 \\ 

\hline

\hline
\end{tabular}
}
\caption{Automatic evaluation results. The best results of standard settings are reported in the \textbf{bold} format. ``ACC'', ``D'', and ``sBL'' are abbreviations of Accuracy, Dist, and Self-BLEU, respectively. ``ACC-CM'' is the average Accuracy of ER, IP, and EX, which are three mechanisms of CM.}
\label{tab-results}
\end{table*}

\subsection{Comparable Methods} 
We compare our method with three groups of representative methods.

\noindent\textbf{Transformer-Based Methods.}
(1) TRS \cite{rashkin-etal-2019-towards} is a vanilla Transformer with MLE loss.
(2) MTRS \cite{rashkin-etal-2019-towards} uses multi-task learning with emotion classification in addition to MLE loss.
(3) MoEL \cite{lin-etal-2019-moel} utilizes different decoders to combine different outputs for each emotion category.
(4) MIME \cite{majumder-etal-2020-mime} applies emotion grouping, emotion mimicry, and stochasticity strategies.
(5) EmpDG \cite{li-etal-2020-empdg} learns emotions and responses based on adversarial learning.
(6) CEM \cite{CEM2021} leverages commonsense to enhance empathetic response generation.

\noindent\textbf{Pre-Trained Language Model-Based Methods.}
(1) TransferTransfo \cite{Wolf2019TransferTransfoAT} is a transfer learning-based GPT-2 \cite{radford2019language} model fine-tuned on \dataset.
(2) BART \cite{lewis-etal-2020-bart} is a pre-trained encoder-decoder Transformer with great success in many seq2seq tasks.

\noindent\textbf{Diffusion Model-Based Method.} 
DiffuSeq \cite{gong2022diffuseq} is proposed as a conditional diffusion language model for seq2seq tasks.

Two more results are provided as references.
Under the Oracle setting, control signals are obtained from golden responses in the test set, which can be regarded as the upper bound of \method. 
Golden responses themselves are also evaluated, which reflects human performance on the task.
More details are listed in Appendix \ref{app-2-2}.

\subsection{Metrics}
\noindent\textbf{Automatic Evaluation.} 
We evaluate the generated responses from four aspects: 
(1) Relevance: \textit{BERTScore} \cite{Zhang2020BERTScoreET} computes a semantic similarity between generated responses and golden references.
\textit{MIScore} is the likelihood of generating a context with the given response, which applies the idea of Maximum Mutual Information (MMI) \cite{li-etal-2016-diversity,Zhang2018GeneratingIA} and indicates whether the generated response is context-related. 
(2) Controllability: We calculate the success rate of empathy expression with multi-grained control signals to validate the controllability of \method. For utterance-level CM and sentence-level IT, we report Accuracy, while for token-level SF, we report F1.
(3) Informativeness: \textit{Dist-n} \cite{li-etal-2016-diversity} calculates the number of distinct n-grams in generated responses. 
\textit{Self-BLEU} \cite{Zhu2018TexygenAB} reflects the difference of all generated responses to a large extent. We calculate the average BLEU-5 overlap between each two generated responses. 
(4) Response Length: \textit{AvgLen} represents the average number of tokens for generated responses. Intuitively, too short text often fails to convey good content.
More details about automatic metrics are shown in Appendix \ref{app-2-3}.

\noindent\textbf{Human Evaluation.}
We evaluate the response quality based on the following aspects: 
(1) \textit{Empathy} reflects whether a response understands the speaker's feeling or situation and responds appropriately.
(2) \textit{Relevance} considers whether a response is relevant to the topic mentioned by the speaker.
(3) \textit{Informativeness} evaluates whether a response provides rich and meaningful information.
More details about the human evaluation guidance are given in Appendix \ref{app-2-4}.

\subsection{Implementation Details}
\method is based on the architecture of BERT-base \cite{Devlin2019BERTPO}.
For diffusion model settings, we adopt the square-root noise schedule \cite{Li-2022-DiffusionLM} and set 2000 diffusion steps in the training and inference process. 
The maximum input length is 128 with WordPiece tokenizer and word embeddings are in the size of 128 with random initialization.
For training settings, we use AdamW optimizer and set the learning rate as 1e-4. 
The batch size and dropout value are set as 128 and $0.1$, respectively.
$\gamma$ in Equation \ref{score-match} equals to $0.2$.
For all comparable methods, we use their official codes with settings that follow the original papers.
For more details, please refer to Appendix \ref{app-2-5}.

%% file: Tex/5_discussion.tex
\section{Results and Discussions}
\subsection{Main Results}

\begin{table}[t]
  \small
  \centering
  \resizebox{0.45\textwidth}{!}{
      \begin{tabular}{l|ccc}
          \hline 
          
          \hline
          \textbf{Method} & \textbf{Empathy} & \textbf{Relevance} & \textbf{Informativeness}\\
          \hline
          \hline
          \textbf{TRS} & 2.96 & 2.49 & 2.31\\
          \textbf{CEM} & 2.84 & 2.69 & 2.75 \\
          \textbf{BART} & 3.04 & 2.94 & 3.92 \\
          \textbf{DiffuSeq} & 2.77 & 2.66 & 3.74\\
          \hline
          \textbf{\method} & \textbf{3.68} & \textbf{3.39} & \textbf{4.63} \\
          \hline
          
          \hline
      \end{tabular}
      }
\caption{Human evaluation results. The Fleiss' kappa \cite{fleiss1973equivalence} of the results is 0.47, indicating a moderate level of agreement.}
\label{table-human}
\end{table}

\begin{table}[h]
  \small
  \centering
  \resizebox{0.45\textwidth}{!}{
  \begin{tabular}{l|cc|cc|c} 
      \hline
      
\hline
\multicolumn{1}{c|}{\multirow{2}{*}{\textbf{Method}}}  & \multicolumn{2}{c|}{\textbf{CM}} & \multicolumn{2}{c|}{\textbf{IT}} & \multicolumn{1}{c}{\textbf{SF}}  \\
\cline{2-6}& \textbf{ACC} ↑ & \textbf{F1} ↑ & \textbf{ACC} ↑ & \textbf{F1} ↑ & \textbf{F1} ↑ \\ 
\hline\hline
          \textbf{\method} & 92.36 & 90.26 & 84.24 & 77.15 & 52.79 \\
          \hline
          \textbf{w/o Mask}& 90.76 & 87.99 & 73.80 & 66.58 & 49.43 \\
          \hline
          \textbf{w/o CM}& \sethlcolor{soulgreen}\hl{89.34} & \sethlcolor{soulgreen}\hl{85.55} & 83.80 & 76.38 & 52.89 \\
          \textbf{w/o IT}& 92.24 & 90.21 & \sethlcolor{soulgreen}\hl{47.92} & \sethlcolor{soulgreen}\hl{41.77} & 52.63 \\
          \textbf{w/o SF}& 89.70 & 86.96 & 83.12 & 74.90 & \sethlcolor{soulgreen}\hl{22.48} \\
          \hline
          
          \hline
      \end{tabular}
      }
\caption{Ablation study on control-range masking and control signals.}
\label{table-control}
\end{table}

\noindent\textbf{Automatic Evaluation Results.}
The overall results are shown in \tableautorefname~\ref{tab-results}.
\method substantially exceeds transformer-based and pre-trained model-based methods on almost all metrics. 
First, the improvement in controllability is significant. 
The high success rate indicates the effectiveness of control-range masking for elaborate token generation and demonstrates the ability of \method to customize responses with desired factors. 
For informativeness, diffusion model-based methods perform the best, and \method is even better than DiffuSeq. 
It has been proven that the diffusion model is a powerful backbone for generating diverse texts.
With the integration of control signals, especially fine-grained signal SF, the meaning of each to-be-generated response token is more specific, thus the final response is more informative.
When considering informativeness values along with MIScore and AvgLen, we can find that those informative responses generated by \method are also context-related and long, which satisfies the demand for proper responses to speakers.
The BERTScore of \method is not the highest, and we think this is reasonable since BERTScore indicates the similarity of generated and golden responses, while \method encourages creativity instead of similarity. 
Besides, the difference between BERTScore and MIScore can justify that the generated responses are both creative and coherent.

\noindent\textbf{Human Evaluation Results.}
Human evaluation results are listed in \tableautorefname~\ref{table-human}.
Our method achieves the highest scores in all aspects, and the greatest improvement is achieved in informativeness, which shows that responses generated by \method are preferred by annotators.
Meanwhile, results of the Oracle setting show that the performance will be further improved when accurate control signals are given, which indicates that obtaining better control signals can be a feasible research topic.

\subsection{Ablation Study}
\noindent\textbf{Ablation on Control-Range Masking.}
To verify the effectiveness of control-range masking, we remove the mask matrix and conduct full self-attention on all input tokens, i.e., input tokens can control or influence the representation of each other. 
As shown in \tableautorefname~\ref{table-control}, the controllability of three signals decreases when the mask is removed (``w/o Mask''), which justifies that our masking strategy is useful for multi-grained control.
Besides, the most significant declines appear at the sentence level, which illustrates that IT has the strongest dependency on the masking strategy. 
We suppose it is because sentence-level signals are not that explicit like token-level signals with word-by-word alignments or utterance-level signals with global modeling in a dialogue session.

\noindent\textbf{Ablation on Control Signals.}
Another question is whether each control signal plays the corresponding role. 
We keep the structure of the control-range mask untouched and remove each signal to validate. 
In detail, we remove the control signal from both the input text and the corresponding row(s) and column(s) in the original mask matrix.
\tableautorefname~\ref{table-control} shows that a success rate decreases when the corresponding control is removed (``w/o CM'', ``w/o IT'', and ``w/o SF''), and the finer the granularity of the control signal, the more the performance declines. 
We can come to the conclusion that each control signal and its control range defined in the mask matrix play an important role in response controllability.

\subsection{Discussions}

\begin{table}[b]
\centering\small
\resizebox{0.45\textwidth}{!}{%
\begin{tabular}{l|r|rr} 
\hline

\hline
\multicolumn{2}{c}{} & \textbf{\method} & \textbf{w/o SF}  \\ 
\hline\hline
\multirow{2}{*}{\textbf{Relevance}} & \textbf{BERTScore ↑} & 52.05     & 51.47   \\ 
& \textbf{MIScore ↓}   & 626.92    & 993.44  \\ 
\hline
\multicolumn{1}{l|}{\multirow{3}{*}{\textbf{Informativeness}}} & \textbf{Dist-1 ↑}        & 2.84      & 1.69    \\ 
\multicolumn{1}{l|}{} & \textbf{Dist-2 ↑}        & 29.26     & 22.83   \\ 
\multicolumn{1}{l|}{} & \textbf{self-BLEU ↓}       & 1.09      & 1.31    \\ 
\hline
\textbf{Length} & \textbf{AvgLen ↑}    & 14.13     & 13.23   \\
\hline

\hline
\end{tabular}
}
\caption{Importance of the fine-grained signal SF.}
\label{table-sfablation}
\end{table}

\noindent\textbf{Analysis on Fine-Grained Signal SF.}
\label{sec-fgcontrol}
Compared with CoMAE \cite{Zheng2021CoMAEAM} which utilizes coarse control signals at the utterance level, we claim that a fine-grained signal is more useful for better empathy expression.
To validate this claim, we remove the fine-grained labels, i.e., token-level SF, to see the performance change.
Results are shown in \tableautorefname~\ref{table-sfablation}. 
Without the token-level control, almost all evaluation metrics decrease in varying degrees.
We conjecture that the token-level guidance gives a direct prompt on the content this token should entail, which greatly narrows the space of acceptable output generation. 

\begin{figure}[t]
    \centering
    \includegraphics[width=0.367\textwidth]{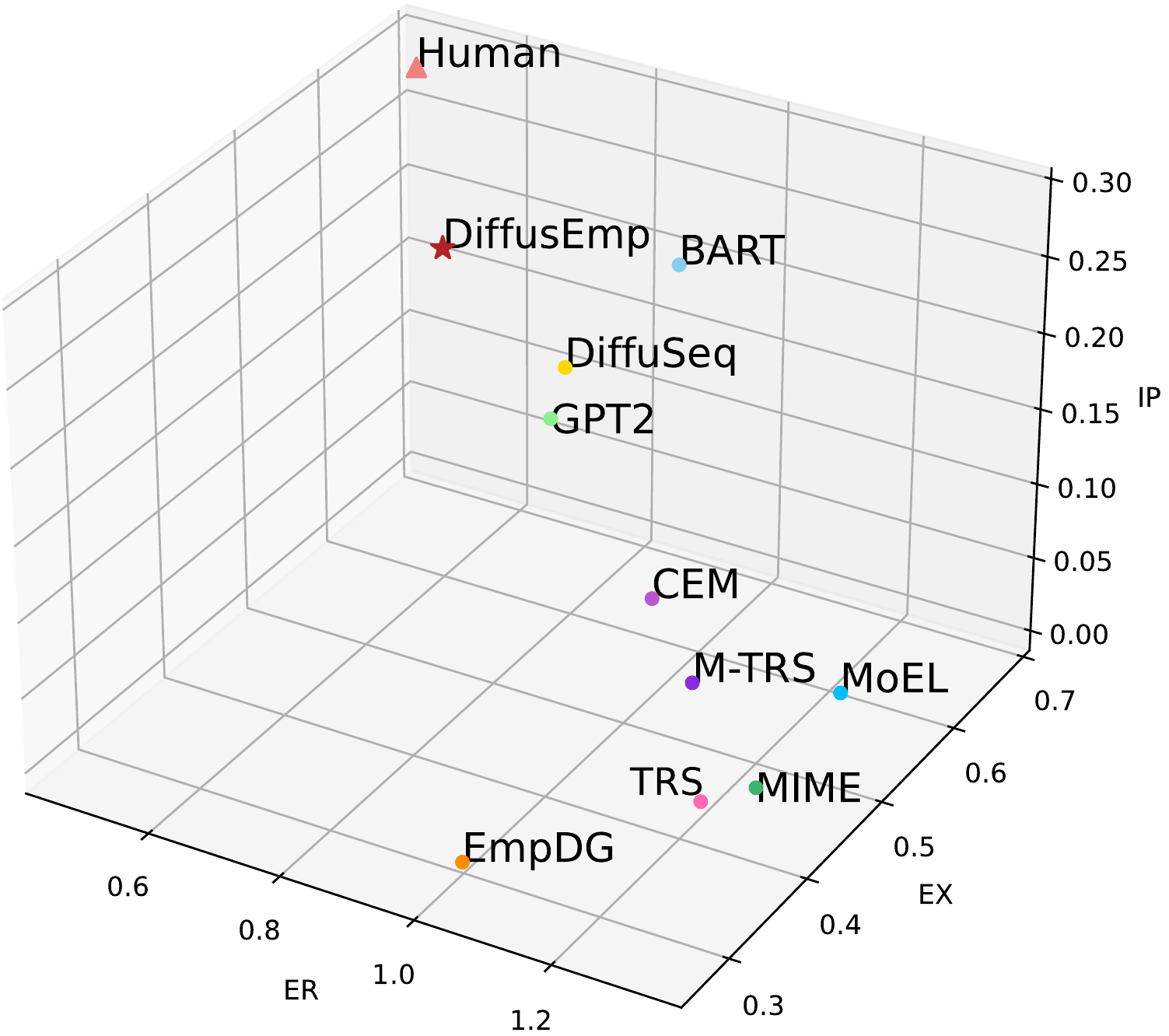}
    \caption{Visualization for CM of different methods.}
    \label{fig-cm}
\end{figure}

\noindent\textbf{Analysis on Coarse-Grained Signal CM.}
Emotional Reaction (ER), Interpretation (IP), and Exploration (EX) are three different high-level mechanisms for empathy expression.
To explore the ways in which different mechanisms express empathy, we score generated responses in these three aspects with RoBERTa-based annotators as mentioned in Section \ref{sec-cf}.
Results are visualized in \figureautorefname~\ref{fig-cm}.
For each method, the average ER, IP, and EX of generated responses on the test set are represented as the coordinate value of a point.
\method is the closest to human responses in distance, indicating that the way our method expresses empathy is the most similar to human beings. 

\begin{table}
\centering
\resizebox{0.47\textwidth}{!}{%
    \begin{tabular}{p{2.5cm}p{7.1cm}}
          \hline 
          
          \hline
          \textbf{Context} &  \textbf{I caught my boyfriend texting his ex.}  \\
          \hline
          \textbf{Golden} & Wow. Dump him and beat him up! \\
          \hline
          \hline
          \textbf{MTRS} & Oh no! What happened? \\
          \hline
          \textbf{MIME} & Oh no, did he get hurt? \\
          \hline
          \textbf{CEM} & What did he do?\\
          \hline
          \textbf{TransferTransfo} & That is terrible! Was he able to get back to you?\\
          \hline
          \textbf{BART} & Oh no! Did you confront him about it?\\
          \hline
          \textbf{DiffuSeq} & Were you hurt? \\
          \hline
          \hline
          \textbf{Candidate A} & Ok $\text{do}^1$ $\text{not}^2$ $\text{feel}^3$ $\text{bad}^4$ $\text{be}$ $\text{happy}^5$ $\text{and}$ $\text{search}^6$ $\text{for}$ $\text{bad}$  $\text{future}^7$ $\text{behalf}$\\
          \hline
          \textbf{Control A} &   \sethlcolor{soulyellow}\hl{\textsc{Emotional\_reaction}}   \sethlcolor{soulblue}\hl{\textsc{Suggesting}} \\
          & \multicolumn{1}{p{7.1cm}}{ \_ \textsc{Intentionally\_act}$^1$ \textsc{No}$^2$ \textsc{Perception\_experience}$^3$ \textsc{Desirability}$^4$ \_ \textsc{Emotion\_directed}$^5$  \_ \textsc{Scrutiny}$^6$ \_ \_ \textsc{Alternatives}$^7$  \_} \\
          \hline
          \textbf{Response A} &  Just do$^1$ not$^2$ feel$^3$ bad$^4$, happy$^5$ to study$^6$ in your future$^7$. \\
          \hline
          \hline
          \textbf{Candidate B} & That could$^1$ be embarrassing, do$^2$ you$^3$ have$^4$ a new$^5$ partner ?$^6$ \\
          \hline
          \textbf{Control B} & \sethlcolor{soulyellow}\hl{\textsc{Exploration}} \sethlcolor{soulblue}\hl{\textsc{Questioning}}  \\
          & \multicolumn{1}{p{7.1cm}}{\_ \textsc{Possibility}$^1$ \_ \_ \_ \textsc{Intentionally\_act}$^2$ \textsc{Pronoun}$^3$ \textsc{Possession}$^4$ \_ \textsc{Age}$^5$ \_ \textsc{?}$^6$} \\
          \hline
          \textbf{Response B} & That could$^1$ be disgusting, do$^2$ you$^3$ have$^4$ a new$^5$ relationship ?$^6$  \\
          \hline
          
          \hline
      \end{tabular}
}
\caption{Case study of \method.}
\label{table-case}
\end{table}

\subsection{Case Study}
\tableautorefname~\ref{table-case} shows the syntactically acceptable examples generated by \method and other comparable methods.
Transformer-based methods tend to generate plain and safe words, lacking a deep understanding of the context. 
In contrast, responses generated by TransferTransfo and BART have more rich information and details. 
All comparable methods tend to respond in general expressions, and even the way to ask questions is also monotonous, which may be due to the large number of such samples in the dataset.
\method responses entail features from both context and guidance. Feelings (\textit{disgusting, don't feel bad}), questions (\textit{new relationship}), and advice (\textit{study for future}) fit the situation of the speaker.
Our framework is also helpful for generating different responses for a given context.
With the support of an emotion-enhanced matching method, multiple response candidates can be returned to further guide response generation with diverse control signals.
Control A and B contain intent \textit{Suggesting} and \textit{Questioning}, respectively.
Thus, \method A aims to give advice while B focuses on asking questions.
More cases are shown in Appendix \ref{app-case}.

%% file: Tex/6_conclusion.tex
\section{Conclusion and Future Work}
We propose \method, a diffusion model-based framework, for empathetic response generation.
To better model multi-dimensional empathy and improve its expression, we utilize multi-grained control signals at utterance, sentence, and token levels.
These control signals are directly extracted from golden responses in the training process, while response candidates obtained from an emotion-enhanced matching method are used as the signal source.
Then we also design a control-range masking strategy and integrate it into the diffusion language model to fulfill elaborate control on the generation of response tokens.
Experimental results on a benchmark dataset \dataset show that our method outperforms competitive baselines in generating more context-related, informative, and empathetic responses. 
Our framework is scalable for more control signal types and can also be extended to other controllable conditional text generation tasks.

In future work, we will extend \method to more empathetic control signals, and improve the performance of annotators and retrieval tools.
Besides, it is interesting to explore \method on various controllable text generation tasks. 

%% file: Tex/7_limitations.tex
\section*{Limitations}
\label{sec-limitation}

The difficulty of obtaining accurately-labeled control signals constrains our results. 
As we report in \tableautorefname~\ref{table-classifier}, the performance of tagging tools can be further improved. However, when the original dataset lacks multi-grained annotations, relying on pre-trained tools is the most feasible solution. 
Considering that control signals come from response candidates in the inference stage, the performance of the context-context matching method is another constraint.
Finally, the drawback of diffusion models also has an impact on our approach.
Despite its high-quality generative performance, the diffusion model has a high requirement for GPU resources and still suffers from slow sampling.
We discuss some attempts to address these limitations in Appendix \ref{app-future}.

%% file: Tex/8_ethics.tex
\section*{Ethics Statement}

The \dataset dataset \cite{rashkin-etal-2019-towards} used to train and evaluate in the paper is collected by crowd-sourcing using the ParlAI platform to interact with Amazon Mechanical Tunk.
Besides, we use \textsc{EmpatheticIntent} \cite{welivita-pu-2020-taxonomy}, \textsc{Reddit} \cite{sharma-etal-2020-computational} and \textsc{FrameNet} \cite{baker-etal-1998-berkeley-framenet} to train tagging tools for control signals. 
All the above datasets are well-established and publicly available. 
Sensitive and personal privacy information have been removed during the dataset construction.
In our human evaluation, participants were fully informed of the purpose of our study and were appropriately compensated.
It is important to clarify that our work is only a study of open-domain dialogue with empathy. We claim that our system does not provide professional psychological counseling. 
In other words, it does not make any treatment recommendations or diagnostic claims.

%% file: Tex/a1_exp_details.tex
\section{Additional Experiment Details}

\subsection{Comparable Methods}
\label{app-2-2}
The following models are chosen as comparable methods and divided into three groups according to their architecture. 

\paragraph{Transformer-Based Methods.}
\begin{itemize}
\item \textbf{TRS}~\cite{rashkin-etal-2019-towards}: A vanilla Transformer with maximum likelihood estimation (MLE) loss.
\item \textbf{MTRS}~\cite{rashkin-etal-2019-towards}: A multi-task model trained with emotion classification loss in addition to MLE loss.
\item \textbf{MoEL}~\cite{lin-etal-2019-moel}: A model using different decoders to generate and combine different outputs for each emotion category.
\item \textbf{MIME}~\cite{majumder-etal-2020-mime}: A model utilizing emotion grouping, emotion mimicry, and stochasticity strategies to generate responses.
\item \textbf{EmpDG}~\cite{li-etal-2020-empdg}: An adversarial model applying two discriminators for interacting with user feedback.
\item \textbf{CEM}~\cite{CEM2021}: A model leverages commonsense as additional information to further enhance empathetic response generation.
\end{itemize}

\paragraph{Pre-Trained Language Model-Based Methods.}
\begin{itemize}
    \item \textbf{TransferTransfo}~\cite{radford2019language, Wolf2019TransferTransfoAT}: A combination of a transfer learning-based training scheme and a high-capacity GPT-2 model which shows strong improvements over end-to-end conversational models.
    \item \textbf{BART}~\cite{lewis-etal-2020-bart}: A pre-trained encoder-decoder Transformer with great success in many seq2seq tasks.
\end{itemize}

\paragraph{Diffusion Model-Based Methods.}
\begin{itemize}
\item \textbf{DiffuSeq}~\cite{gong2022diffuseq}: A diffusion model proposed as a conditional language model and trained end-to-end in a classifier-free manner. It is designed for sequence-to-sequence text generation tasks.
\end{itemize}

Noticed that we did not use Diffusion-LM~\cite{Li-2022-DiffusionLM} as a baseline because it is incompatible with the sequence-to-sequence task setting.
We provide the result of \textit{oracle setting} as a reference. Under the standard setting, the attributes are not given and need to be predicted from the retrieve-based methods, and we focus on evaluating the response quality. Under the oracle setting, the true attributes from the ground truth response are provided, so it can be considered as the theoretical upper limit performance of \method.

\subsection{Automatic Evaluation}
\label{app-2-3}
We evaluate the generated empathetic responses from the following four aspects: relevance, controllability, informativeness, and response length.

\paragraph{Relevance.}
We use \textit{BertScore} and the \textit{MIScore} of response to evaluate relevance.
\begin{itemize}
\item \textbf{BertScore}~\cite{Zhang2020BERTScoreET}:
BertScore computes a similarity score using contextual embeddings for each token in the candidate sentence with each token in the reference sentence. We use \textit{deberta-large-mnli} to calculate the BertScore.

\item \textbf{MIScore}:
A good response should be informative and relevant to the context. When given the response, it should have the ability to infer its context, while a safe response is generic and can be used in any context, so it is hard to infer the context. From this perspective, we use the idea of \textit{Maximum Mutual Information (MMI)}~\cite{li-etal-2016-diversity, Zhang2018GeneratingIA}. The idea of MIScore is employing a pre-trained backward model to predict context sentences from given responses, i.e., $P(\text{Context}|\text{Response})$. 
Intuitively, MIScore encourages the model to generate responses that are more specific to the context, while generic responses are largely less preferred, since they can be used in any case. 
We calculate MIScore according to the following equation:
$$\exp(-\frac{1}{m}\sum^m_{t=1}\log P(x_t|y_1,\dots,y_n,x_{<t}),$$
where $m$ and $n$ are the numbers of tokens in the context and response respectively. 
It is implemented with a reverse 345M DialoGPT~\cite{zhang2019dialogpt}, which is a fine-tuned GPT-2 ~\cite{radford2019language} with the training objective to predict the context from the response.
\end{itemize}

\paragraph{Controllability.}
We calculate the attribute control accuracy success rate to validate the controllability of models. For session-level CM and sentence-level IT, we report accuracy. For token-level SF, we report F1.

\paragraph{Informativeness.}
We use \textit{Distinct n-gram}~\cite{li-etal-2016-diversity} and \textit{self-BLEU}~\cite{Zhu2018TexygenAB} to evaluate informativeness.
\begin{itemize}
    \item\textbf{Distinct n-gram}~\cite{li-etal-2016-diversity}: Distinct n-gram calculates the number of distinct n-grams in generated responses. The value is scaled by the total number of generated tokens to avoid favoring long sentences.
    \item\textbf{Self-BLEU}~\cite{Zhu2018TexygenAB}: Self-BLEU regards one sentence as a hypothesis and the others as a reference, we can calculate the BLEU score for every generated sentence, and define the average BLEU score to be the Self-BLEU of the document. 
\end{itemize}

\paragraph{Response Length.}
\begin{itemize}
    \item \textbf{Average Length}~\cite{Singh2016RankingSF}:
    The length of the response text is also used as a quality indicator when comparing different model generations since shorter texts usually contain less information.
\end{itemize}

It is noteworthy that open-domain dialogue and controllable text generation contain a great deal of creativity.
When a sentence is forced to remain identical to a fixed standard sentence, such evaluation metrics may unfairly penalize creative texts, notwithstanding they are capable of responding to the given context.
As a result, instead of comparing the word overlap between generated responses and standard responses, we give the metric values of standard responses as a reference.

\subsection{Human Evaluation}
\label{app-2-4}
Quantitative automatic metrics are straightforward to compare, but they may be less effective at reflecting overall levels of empathy. 
Human judgment is necessary for an open-domain dialogue system~\cite{liu-etal-2016-evaluate}. 

We recruit three third-party graduate researchers (average age 23.3) to analyze the results of various models. We acquired permission for their participation and paid them in accordance with local hourly wages.
The response quality of all models is evaluated in terms of the following three aspects: Empathy, Relevance, and Informativeness.
We randomly sample 100 dialogues and corresponding generated responses for different models and then ask three professional annotators to give each response a rating score from the following aspects.
\begin{itemize}
    \item \textit{Empathy} reflects whether the listener understands the feeling of the speaker and responds appropriately.
    \item \textit{Relevance} considers how the content of the reply is relevant to the topic mentioned by the speaker.
    \item \textit{Informativeness} evaluates grammar correctness and readability.
\end{itemize}

\begin{figure*}[t]
    \centering
    \includegraphics[width=0.6\textwidth]{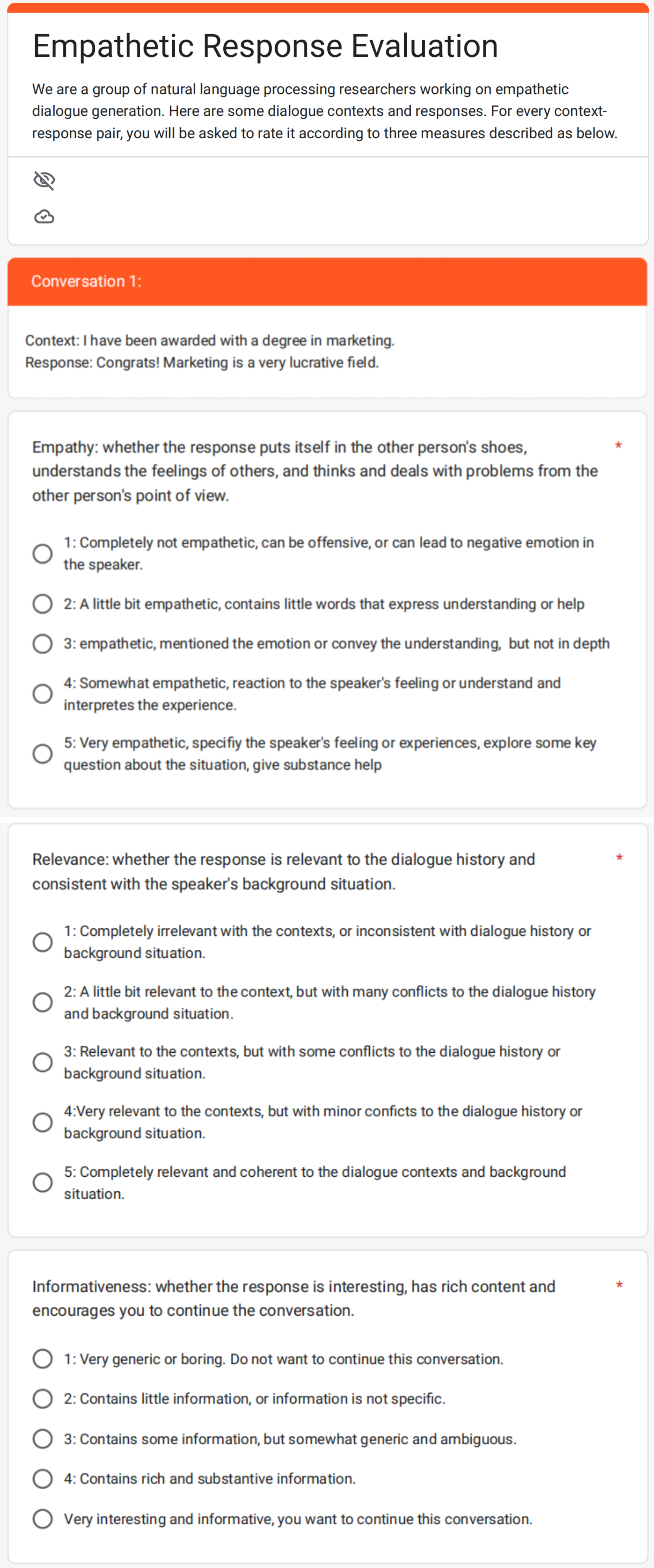}
    \caption{An example of the survey for our human evaluation.}
    \label{cm-humanevalguide}
\end{figure*}

The specific instruction given to them for the evaluation is shown in~\figureautorefname~\ref{cm-humanevalguide}.
Each aspect is on a scale of 1 to 5, in which 1 is ``unacceptable'' and 5 is ``excellent performance''.

Besides, We conduct an A/B test to directly compare our method with other baselines. 
Another 100 dialogues are randomly sampled from each model. 
Three annotators are given generated responses from either our method or baselines in random order and are asked to choose a better one. 
They can either choose one of the responses or select ``Tie'' when the quality of provided options is hard to access.

\subsection{Implementation Details}
\label{app-2-5}
Our \method calculates diffusion model parameters with a BERT-base~\cite{Devlin2019BERTPO} architecture with 12 layers and 80M parameters.
For diffusion settings, we set 2000 diffusion steps in both the training stage and the inference stage. We adopt the square root noise schedule. The max input length is 128, the dimensions of word embedding and time embedding are all 128, and the embedding is randomly initialized\footnote {We also attempt the initialization with pre-trained bert-base-uncased vocabulary but the result is poor.}.
For training settings, we use AdamW optimizer and set the learning rate as 1e-4, dropout as $0.1$. We set gradient clipping to $-1.0$. 
$\gamma$ equals to $0.2$.
We use WordPiece tokenizer\footnote{Firstly we try to build vocabulary for our own dataset but find it heavily suffers from the out-of-vocabulary problem.}. The batch size is 128 and the micro-batch size is 64. 
For all baseline models, we use their official codes to implement and keep the settings in the original paper.

%% file: Tex/a2_future.tex
\section{Future Work}
\label{app-future}

The limitations of our work have been mentioned in Section~\ref{sec-limitation}. Here, we propose some attempts to overcome these limitations.

\paragraph{Control Signals.}
In the acquisition of control signals, there are two main constraints for performance, including (1) the accuracy of control signals and (2) the suitability of retrieval results in the testing step.

With regard to (1), the results of the oracle setting demonstrate that our framework has a high ceiling when ground-true control signals are given.
Therefore, we have tried to enhance robustness by noising the control factors. Noising methods contain adding, removing, and replacing random control tokens. 
However, experimental results show that noising methods compromise the success rate of control, which is contrary to the motivation of this work. 
In the future, this approach can be tried to further improve language quality in scenarios where the demand for controllability is weak.

With respect to (2), we focus on the performance of the retrieval model in the inference stage. The control signals straightforwardly come from the retrieved responses. In this paper, we have proposed a task-specific design that combines semantic and emotional similarity to retrieve but it is still simple compared to those SOTA dialogue response selection models. 
In future work, it is meaningful to replace our retrieval model with more powerful response selection methods.

As an advantage of \method, both the annotating taggers and the retrieval model are orthogonal to empathetic response generation.
It is easy for followers to employ higher-performance response selection models and attribute annotating taggers to empower the \method.

\paragraph{Diffusion Models.}
Finally, the diffusion model requires a lot of GPU computational resources and is slow when inference, which limits its application.  
There are many attempts to reduce the computational resources~\cite{Rombach2021HighResolutionIS} required by the diffusion model as well as to speed up the process~\cite{Vahdat2021ScorebasedGM} and inference~\cite{Song2020DenoisingDI, Bao2022AnalyticDPMAA}. Theoretically, the relevant improvements would also have an enhancing effect on our framework and would be helpful for spreading the diffusion model to the NLP community.

%% file: Tex/a3_cases.tex
\section{Case Study}
\label{app-case}
We give more responses generated by \method in \tableautorefname~\ref{app-case-1} and \tableautorefname~\ref{app-case-2}.
\begin{table*}
    \centering\small
    \resizebox{0.98\textwidth}{!}{%
    \begin{tabular}{lp{12cm}} 
    \toprule[1.5pt]
    \multicolumn{2}{c}{\textbf{Case 1}}    \\ 
    \midrule
    \textbf{Context}   & \textbf{I feel lonely sometimes cause I live in a house with 5 girls and all of them have boyfriends and sometimes they are all hanging out with their significant others and I feel lonely.}\\
    \midrule
    \textbf{Gold}   & Awww, That sucks man. Hopefully they don't say mean things to you about being lonely.                 \\
    \hline\hline
    \textbf{TRS} & I am sorry to hear that. I am sorry to hear that. \\
          \midrule
          \textbf{Multi-TRS} & I am sorry to hear that. I am sorry to hear that. \\
          \midrule
          \textbf{MoEL} & I am sorry to hear that. have you tried talking to them?\\
          \midrule
          \textbf{MIME} & I am sorry to hear that. I am sorry to hear that. I hope you are able to do that. \\
          \midrule
          \textbf{EmpDG} & That is a shame. I am sorry to hear that.\\
          \midrule
          \textbf{CEM} & I am so sorry to hear that. I am sure you will be fine. \\
          \midrule
          \textbf{GPT2-FT} & Why were they so angry?\\
          \midrule
          \textbf{BART-FT} & I'm sorry to hear that. Have you tried joining any clubs?\\
          \midrule
          \textbf{DiffuSeq} & Why can not you too \\
          \hline\hline
    \textbf{Candidate A} & I see. I'm sorry you're home alone. I get lonely myself when you're used to someone around. are you going to advertise for a new roommate? \\
    \midrule
    \textbf{Control A}  & \textsc{Emotional\_reaction Exploration Interpretation Acknowledging Sympathizing Neutral Questioning Pronoun Perception\_experience \_ Pronoun \_ \_ Pronoun \_ Buildings \_ \_ Pronoun Transition\_to\_state \_ \_ When Pronoun \_ Using \_ People \_ \_ \_ Pronoun Motion \_ \_ \_ \_ Age \_ ?}\\
    \midrule
    \textbf{Response A} & I see. I'm sorry you're home alone. I got to lonely when I've used to noone with. are you going to find a brand new friend? \\
    \midrule[1pt]
    \textbf{Candidate B} &  Maybe you could invite a friend over for a movie?   \\
    \midrule
    \textbf{Control B}  &  \textsc{Exploration Suggesting \_ Pronoun Possibility Request \_ Personal \_ Relationship \_ \_ \_ \_ ?}\\
    \midrule
    \textbf{Response B} &  maybe you could call a friend to talk with her? \\ 
    \midrule[1pt]
    \textbf{Candidate C} &  Oh no really, that really sucks to hear. Why were you not able to go?  \\
    \midrule
    \textbf{Control C}  &  \textsc{Emotional\_reaction Exploration  Acknowledging Questioning \_ No \_ \_ \_ \_ \_ \_ Perception\_experience \_ Why \_ Pronoun No Capability \_ Motion ? }\\
    \midrule
    \textbf{Response C} & Oh no, that is really amazing to hear. Why were you not able to move? \\ 
    \bottomrule[1.5pt]
    \end{tabular}
    }
    \caption{Cases generated by \method with different \cs.}
    \label{app-case-1}
\end{table*}

\begin{table*}
    \centering\small
    \resizebox{0.98\textwidth}{!}{%
    \begin{tabular}{lp{12cm}} 
    \toprule[1.5pt]
    \multicolumn{2}{c}{\textbf{Case 2}}    \\ 
    \midrule
    \textbf{Context} & \textbf{Today traffic was horrible and was so frustrating!} \\
    \midrule
    \textbf{Gold} & I hate traffic too, it makes me angry\\
      \hline\hline
      \textbf{TRS} & Oh no! What happened? \\
      \midrule
      \textbf{Multi-TRS} & Oh no! I hate when that happens. Did you get hurt?\\
      \midrule
      \textbf{MoEL} & I am sorry to hear that. What happened?\\
      \midrule
      \textbf{MIME} & I am sorry to hear that. What happened?\\
      \midrule
      \textbf{EmpDG} & I hate when that happens.\\
      \midrule
      \textbf{CEM} & What happened?\\
      \midrule
      \textbf{GPT2-FT} & Why were they so angry?\\
      \midrule
      \textbf{BART-FT} & I hate traffic too. I hate it when it's so bad.\\
      \midrule
      \textbf{DiffuSeq} & Was it hurt you ? \\
      \hline\hline
    \textbf{Candidate A} & I completely understand. I get pretty aggravated myself driving home from work everynight.\\\midrule
    \textbf{Control A}  &  \textsc{Emotional\_reaction Interpretation Agreeing Agreeing Pronoun \_ Awareness \_ Pronoun \_ \_ Experiencer\_obj \_ Subjective\_influence Buildings \_ Work \_ \_} \\
    \midrule
    \textbf{Response A} & I completely understand. I have been tired to drive home from work everyday.  \\ 
    \midrule
    \textbf{Candidate B} & Yes! Whats even worse is when other people don't pay attention in bad traffic! \\\midrule
    \textbf{Control B}  &  \textsc{Interpretation Suggesting Questioning Yes \_ \_ \_ \_ \_ \_ Increment People \_ No Commerce\_Pay Attention \_ Desirability \_ \_ }                \\
    \midrule
    \textbf{Response B} & Yes! Traffics is the worst but other people don't pay attention to bad thing. \\
    \midrule
    \textbf{Candidate C} &  Yes, the cable company is infuriating. do they eventually help you though?\\\midrule
    \textbf{Control C}  &  \textsc{Exploration Neutral Questioning Yes \_ \_ \_ Businesses \_ \_ \_ Intentionally\_act Pronoun Time\_vector Assistance Pronoun Concessive?}                \\
    \midrule
    \textbf{Response C} & Yes, the bus company was annoying. Did they already help you out?                  \\
    \bottomrule[1.5pt]
    \end{tabular}}
    \caption{Cases generated by \method with different \cs.}
    \label{app-case-2}
\end{table*}